# ACCELERATING THE ANT COLONY OPTIMIZATION BY SMART ANTS, USING GENETIC OPERATOR

Hassan Ismkhan

Department of Computer Engineering, University of Bonab, Bonab, East Azerbaijan, Iran
H.Ismkhan@bonabu.ac.ir

## ABSTRACT

*This paper research review Ant colony optimization (ACO) and Genetic Algorithm (GA), both are two powerful meta-heuristics. This paper explains some major defects of these two algorithm at first then proposes a new model for ACO in which, artificial ants use a quick genetic operator and accelerate their actions in selecting next state.*

*Experimental results show that proposed hybrid algorithm is effective and its performance including speed and accuracy beats other version.*

## 1. INTRODUCTION

The versions of ACO have been successfully applied to type of combinatorial optimization problem same as scheduling problems [5], routing problems [6] and knapsack problem [9]. The ACO is also successful in dealing with continuous optimization problems [11] [12]. However time complexity of ACO is $\Omega(N^2)$ which is too high [10] and prevents applicability of ACO for large scale problems. In other hand, versions of Genetic Algorithm (GA) which is one of the most important type of meta-heuristics, have been also applied to combinatorial and continues optimization problems. The types of GA have solved scheduling problems [1], knapsack problems [14] and routing problems [15] which all are combinatorial optimization problems. GAs also have been successfully applied to continuous optimization problems such as function optimization [16] [17].

Although GA and ACO are successful meta-heuristics, both have their own bottlenecks. Considering defect points of GA and ACO will help us to design effective frameworks.

GA has powerful population-based search engine which can explore new regions of answer area, instead its exploitation ability dramatically comes down when appropriate operators is not used in body of GA. It should be considered that designing new operators for GA needs more invention. Operators should have enough speed and should increase exploitation ability of GA, where population of GA grant exploration ability of GA.

ACO usually uses small group of artificial ants to make solutions. In addition, ACO uses additional information in its baseline algorithm which is known as pheromone trails. Pheromone update by ants and help ants to find better solutions. Although using pheromone causes appropriate level of exploitation, these additional information decrease exploration ability of algorithm and led to local trap problem.

There are some research attempt to use and merge benefits of both GA and ACO [7]. This paper also proposes effective model of merging ACO and GA which has appropriate speed and accuracy. Therefore the rest of paper scheduled as, next section reviews GA and ACO briefly, section 3 propose our method, section 4 puts forward experimental results and finally section 5 summarizes paper.

## 2. GA AND ACO, BRIEF REVIEW OF THEIR ALGORITHMS

The case study of this research is Travelling Salesman Problem (TSP). TSP is finding minimum Hamilton tour in withed graph. Also in this section, both GA and ACO are explained by using TSP.

### 2.1. ACO

Ant System (AS) is first version of ACO. In AS, in each step, m artificial ants locate on m different start nodes and begin to construct their tours in n sub-steps, where n is the number of nodes in graph. Each ant to select next node uses a rule which is named transition rule and is defined as below:

$$p_k(r,s) = \begin{cases} \frac{[\tau(r,s)]^\alpha \cdot [\eta(r,s)]^\beta}{\sum_{u \in J_k(r)}[\tau(r,u)]^\alpha \cdot [\eta(r,s)]^\beta} & \text{if } s \in J_k(r) \\ 0 & otherwise \end{cases} \quad (1)$$

Where $p_k(r,s)$ is probability of a choosing node *s* after node *r* by ant *k*, $J_k(r)$ is the set of unvisited nodes of ant *k*, $\eta(r,u) = 1/distance(r,u)$ and $\tau(r,s)$ is pheromone amount of *rs*. The β and α are parameters. When each ant completes its tour, tries to update pheromone value by below equation which is named pheromone update rule.

$$\tau(r,s) \leftarrow (1-\alpha).\tau(r,s) + \sum_{k=1}^{m}\Delta\tau_k(r,s) \quad (2)$$

Where $\Delta\tau_k(r,s) = \begin{cases} \frac{1}{tour\ length\ of\ ant\ k}, & if\ edge\ (r,s) is\ in\ ant\ k\ tour \\ 0, & otherwise \end{cases}$

Ant Colony System (ACS) is another versions of ACO which add another rule in which ants after each sub-step update pheromone locally when select their next node. This rule is named local pheromone update rule and implemented by below equation.

$$\tau(r,s) \leftarrow (1-\rho).\tau(r,s) + \rho.\tau_0 \quad (3)$$

In ACS transition rule uses below eqution:

$$S = \begin{cases} \arg\max_{u \in J_k(r)}\{[\tau(r,u)]^\alpha \cdot [\eta(r,u)]^\beta\}, & \text{if } q \leq q_0 \\ use\ (1) & , otherwise \end{cases} \quad (4)$$

Max-Min Ant System is another version of ACO in which pheromone value are bounded between a max and min value. Both methodologies in ACS and MMAS help ACO to escape from local optima traps [13]. The baseline algorithm for ACO is as figure 1.

```
For I = 1 to MAX-Number-of-Iteration

    Locate m ants on m different nodes

   For J = 1 to N (N is the number of nodes in graph)

      For each ant

         Select next node according to (4)

         Update pheromone locally by (2)

      End For

   Update pheromone globally by (3)

End For
```

Figure 1. ACO algorithm

## 2.2. GA

GA usually starts with population of random solutions. In each step, new solutions are generated by operators. These operators select and take one or more solutions from population and produce new solution according to selected solutions.

There are two types of operator for GA: mutation operator and crossover operator. Mutation operator, takes a solution and changes it. This operator helps GA to prevent increasing similar solutions in population. This is a benefit, because similar solutions in population causes premature convergence and local optima problems. The Double-Bridge is one of the most famous mutation for TSP solver GAs. Figure 2 shows Double-Bridge operator.

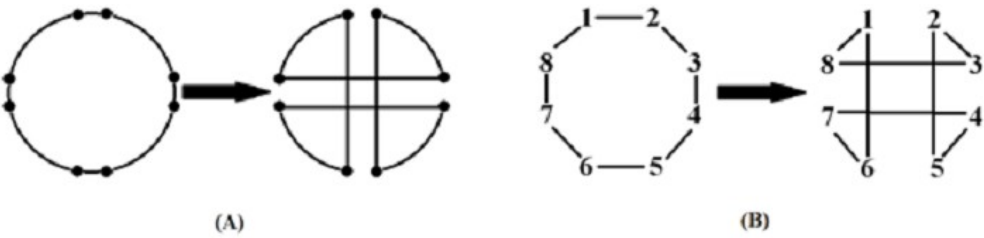

Figure 2. The Double-Bridge operator: Part (A) Shows how Double-Bridge works, Part (B) Shows an example for Double-Bridge

In other hand, crossover operator takes two solutions usually are named parent, and produces one or two solutions according to (similar to) parents which are named children or offspring. Paper [3] reviews some of recent crossover operators. There are some accurate crossover operators same as EAX [8] but its time complexity is o($n^2$) which is too slow. Here we point to Improved Greedy Crossover which has o(n) time complexity [7]. To explain IGX, please consider a graph with 8 nodes that its edges cost are in matrix of figure 4. Figure 5 shows how IGX constructs child according to the two parents as 1-6-8-4-5-7-3-2 and 6-2-4-8-5-1-7-3.

|   | 1 | 2 | 3 | 4 | 5 | 6 | 7 | 8 |
|---|---|---|---|---|---|---|---|---|
| 1 | 0 | 12 | 19 | 31 | 22 | 17 | 23 | 12 |
| 2 | 12 | 0 | 15 | 37 | 21 | 28 | 35 | 22 |
| 3 | 19 | 15 | 0 | 50 | 36 | 35 | 35 | 21 |
| 4 | 31 | 37 | 50 | 0 | 20 | 21 | 37 | 38 |
| 5 | 22 | 21 | 36 | 20 | 0 | 25 | 40 | 33 |
| 6 | 17 | 28 | 35 | 21 | 25 | 0 | 16 | 18 |
| 7 | 23 | 35 | 35 | 37 | 40 | 16 | 0 | 14 |
| 8 | 12 | 22 | 21 | 38 | 33 | 18 | 14 | 0 |

Figure 4. The cost matrix of a graph with 8 nodes

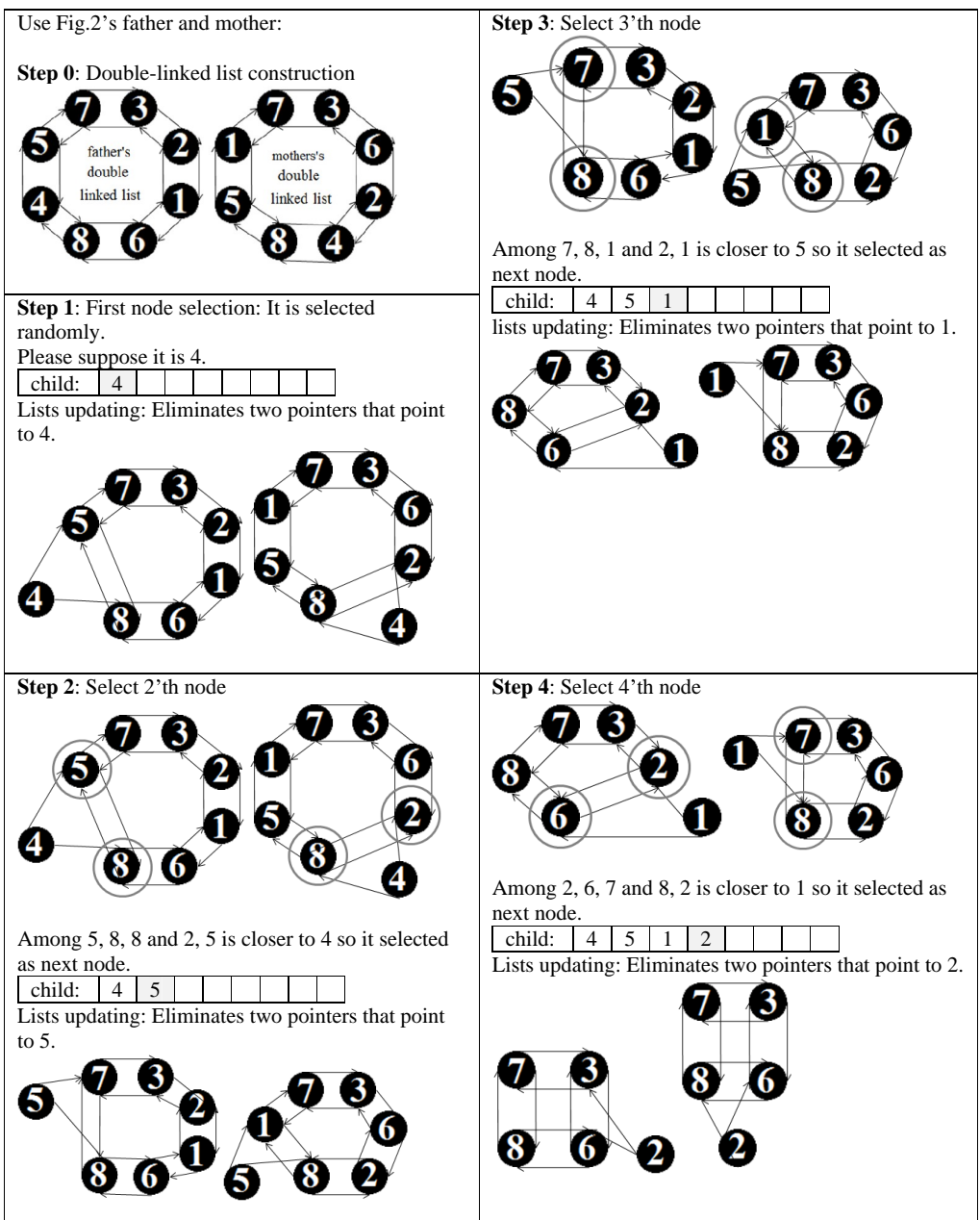

Figure 5. IGX [2]

GA, itself has two important types: generational GA and steady state GA. In generational GA, in each generation, new produced solutions (which are produced by mutation or crossover operators) are added to population and after some generations, the size of population normalized by eliminating solutions with low fitness. In steady state GA, in each generation, new solution is replaced by one of solution in population. Figure 6 shows baseline algorithm of both generational and steady state GA.

| Initial population; | Initial population; |
|---|---|
| **For** gen-num = 1 to Max-Gen | **For** gen-num = 1 to Max-Gen |
|   **while**(some conditions) |   select individuals to apply crossover; |
|     select individuals to apply crossover; |   apply crossover and generate individual(s); |
|     apply crossover and generate individual(s); |   mutate new individual(s); |
|     if(random number < MR) |   Replace new individual(s) with some ones from population |
|      mutate new individual(s); | **End For** |
|     Add new individuals to population; | |
|   **End While** | |
|   Remove extra individuals from population; | |
| **End For** | |
| Generational GA | Steady state GA |

Figure 6. The two types of GA

## 3. PROPOSED SMART ANT

As you see in previous section, IGX selects one edge with lowest cost among four edges. Proposed smart ant are designed same as IGX, instead probing costs of four edges, probes results of transition rule of ACS which stated in equation 5.

$$S = \begin{cases} \arg\max_{u \in PC}\{[\tau(r,u)]^\alpha \cdot [\eta(r,u)]^\beta\}, & \text{if } q \leq q_0 \\ \text{use (6)} & \text{, otherwise} \end{cases} \quad (5)$$

$$p_k(r,s) = \begin{cases} \frac{[\tau(r,s)]^\alpha \cdot [\eta(r,s)]^\beta}{\sum_{u \in PC}[\tau(r,u)]^\alpha \cdot [\eta(r,s)]^\beta} & \text{if } s \in PC \\ 0 & otherwise \end{cases} \quad (6)$$

Where all notations in (5) and (6) are same as (1) and (4) and PC is set of all four nodes which are neighbour of current node in both parents.

Please consider that proposed hybrid algorithm consider a population of solutions and in each step, smart ant which operates as IGX, takes two solutions from population, and produce new child. In this process, smart ant considers equation 6 instead of cost of edges.

```
Initial population;

For gen-num = 1 to Max-Gen

    Assign two solution for each ant

    For each ant

        For I = 1 to N

            Select next node according to 5

            Update pheromone locally

        End For

        //Solution of this ant is now completed
        Apply 2-Opt [13] local search on solution of this ant
        Update pheromone globally by equation 3

End For
```

Figure 6. Proposed hybrid ACO with smart ants

## 4. EXPERIMENTAL RESULTS

To probe performance of Hybrid-ACO-with-Smart-Ant (proposed method), call it HACO-SA, we set up an experiment in which HACO-SA competed with MMAS to solve instance selected from TSPLIB [5]. We ran this experiment on 2.40 GHz Intel CPU with 1 GB of random access memory. All algorithm were implemented by c++ programming language. Experimental results show that accuracy of HACO-SA is as well as MMAS, however speed of HACO-SA is more than MMAS. Table 1 shows average (the number of runs was 30) cost of solutions gained by each algorithm. As can been seen in table I, HACO-SA even has better solutions in average and for last four instances completely overcomes MMAS.

Table 1. Average costs, computed by each algorithm per each instance.

| Average cost | MMAS | HACO-SA |
|---|---|---|
| eil51 | 426 | 426 |
| eil76 | 538 | 538 |
| kroA100 | **21282** | 21283.5 |
| lin105 | **14379** | 14410.29 |
| d198 | **15883.12** | 15983.5 |
| lin318 | 42267.5 | **42219.33** |
| pcb442 | 51010.233 | **50907.15** |
| att532 | 27837.11 | **27761.3** |
| rat783 | 8899.3 | **8851.21** |

Table 2 shows speed of HACO-SA is better than MMAS for last four instances. Also for lin105, HACO-SA is quick than MMAS.

Table 2. Average time

| Average time | MMAS | HACO-SA |
|---|---|---|
| eil51 | 1.4887767 | 1.759 |
| eil76 | 1.8801767 | 2.27605 |
| kroA100 | 3.859565 | 4.1837 |
| lin105 | 4.4618017 | 3.16685 |
| d198 | 9.13653 | 10.13775 |
| lin318 | 19.974383 | 15.9019 |
| pcb442 | 16.971528 | 16.46665 |
| att532 | 46.208823 | 34.2132 |
| rat783 | 46.455887 | 36.1391 |

## 5. CONCLUSIONS

This paper reviews GA and ACO algorithms which both are two important types of meta-heuristics. Although these meta-heuristics are successful in dealing with optimization problems, both face with some critical defects. These paper lists some of these problems and explains how can eliminate these by utilizing good properties of these two methods in a hybrid way.

According to these, this research propose a hybrid version of ACO in which ants act as greedy genetic operator. Experimental results show that proposed method hybrid method with these smart ants has better performance in both factors of speed and accuracy.

**Authors**

Hassan Ismkhan is instructor at the University of Bonab.